\documentclass[journal,twoside,web]{ieeecolor}
\usepackage{tmi}
\usepackage{cite}
\usepackage{amsmath,amssymb,amsfonts}
\usepackage{algorithmic}
\usepackage{graphicx}
\usepackage{textcomp}
\usepackage{float} 
\usepackage{subfigure}
\usepackage{booktabs}
\usepackage{bm}
\usepackage{algorithm}
\usepackage{pgfplots}
\usepackage{mathtools}
\usepackage{url}
\usepackage{cite}
\usepackage{color}

\newcommand{\etal}{\textit{et al.}}
\definecolor{ForestGreen}{RGB}{34,139,34}

\def\BibTeX{{\rm B\kern-.05em{\sc i\kern-.025em b}\kern-.08em
    T\kern-.1667em\lower.7ex\hbox{E}\kern-.125emX}}
\begin{document}
\title{GraVIS: Grouping Augmented Views from Independent Sources for Dermatology Analysis}
\author{Hong-Yu Zhou, \IEEEmembership{Member, IEEE}, Chixiang Lu, Liansheng Wang, \IEEEmembership{Member, IEEE}, \\
and Yizhou Yu, \IEEEmembership{Fellow, IEEE}
\thanks{(\emph{Corresponding author: Yizhou Yu.})}
\thanks{Hong-Yu Zhou, Chixiang Lu and Yizhou Yu are with the Department of Computer Science, The University of Hong Kong, Pokfulam, Hong Kong (e-mail: whuzhouhongyu@gmail.com, luchixiang@gmail.com, yizhouy@acm.org).}
\thanks{Liansheng Wang is with the Department of Computer Science, Xiamen University, Siming District, Xiamen, Fujian Province, P.R. China (e-mail: lswang@xmu.edu.cn).}
\thanks{\emph{First two authors contributed equally.}}
}
\maketitle
\begin{abstract}
Self-supervised representation learning has been extremely successful in medical image analysis, as it requires no human annotations to provide transferable representations for downstream tasks. Recent self-supervised learning methods are dominated by noise-contrastive estimation (NCE, also known as contrastive learning), which aims to learn invariant visual representations by contrasting one homogeneous image pair with a large number of heterogeneous image pairs in each training step. Nonetheless, NCE-based approaches still suffer from one major problem that is one homogeneous pair is not enough to extract robust and invariant semantic information. Inspired by the archetypical triplet loss, we propose GraVIS, which is specifically optimized for learning self-supervised features from dermatology images, to group homogeneous dermatology images while separating heterogeneous ones. In addition, a hardness-aware attention is introduced and incorporated to address the importance of homogeneous image views with similar appearance instead of those dissimilar homogeneous ones. GraVIS significantly outperforms its transfer learning and self-supervised learning counterparts in both lesion segmentation and disease classification tasks, sometimes by 5 percents under extremely limited supervision. More importantly, when equipped with the pre-trained weights provided by GraVIS, a single model could achieve better results than winners that heavily rely on ensemble strategies in the well-known ISIC 2017 challenge. Code is available at \url{https://bit.ly/3xiFyjx}.

\end{abstract}

\section{Introduction}
\begin{figure}[t]
    \centering
    \includegraphics[width=0.6\columnwidth]{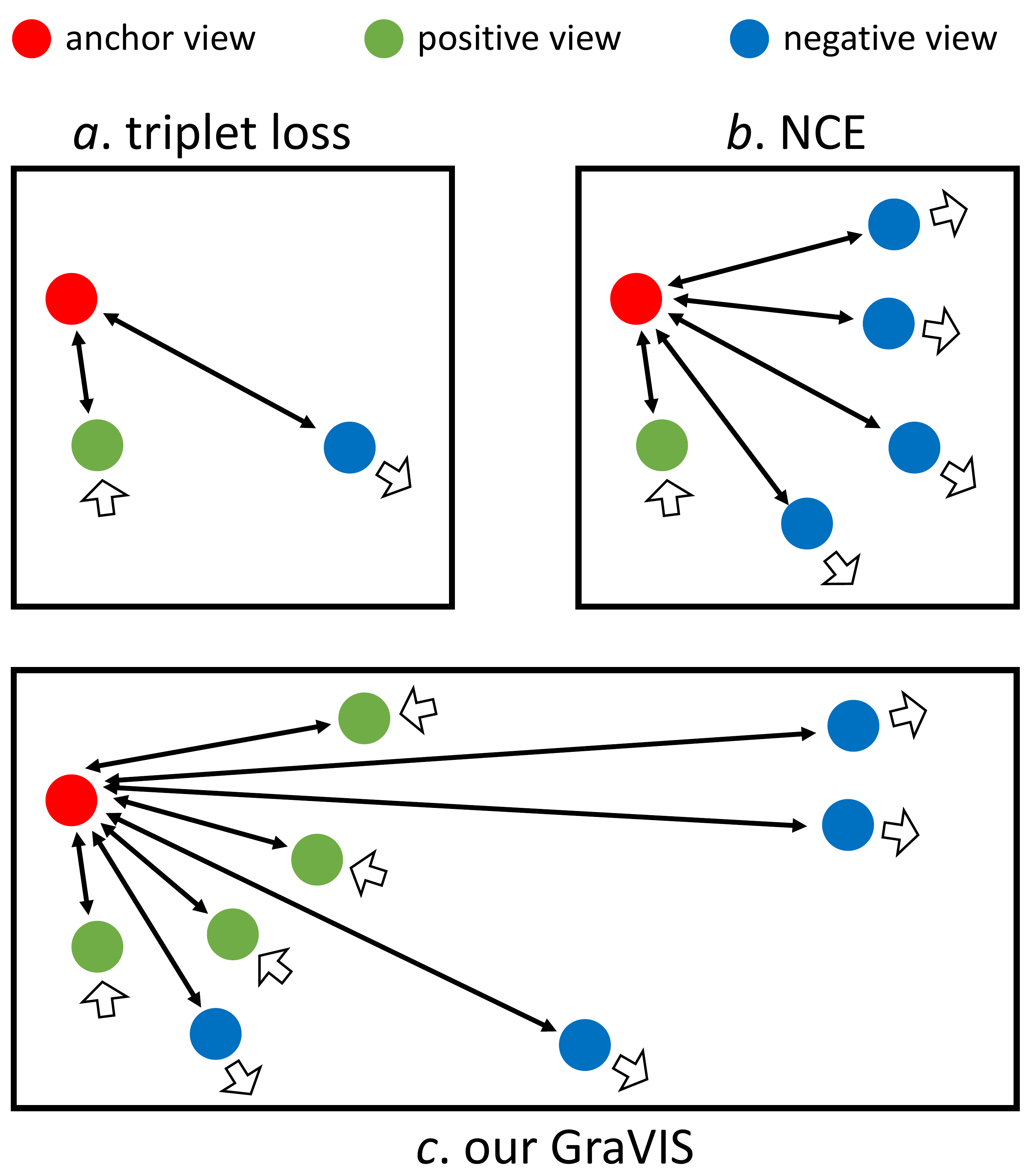}
    \caption{Comparison with the archetypical triplet loss (\emph{a}) and noise-contrastive estimation (\emph{b}. NCE). The hollow arrow denotes the direction to be pushed, where positive views are pushed towards the anchor view and negative views are pushed away. Here, one anchor view and one positive view construct one homogeneous pair. Accordingly, one heterogeneous pair consists of one anchor view and one negative view. To generate positive views, we randomly augment the same image as used to generate the anchor view, while negative views are produced by augmenting different images.}
    \label{fig:intro}
\end{figure}

Considering the protection of patients' privacy and limitation of accessible expertise, self-supervised learning (SSL) \cite{zhou2021models,zhou2020comparing,azizi2021big,chen2020simple,he2020momentum,zhou2021preservational,zhou2017sunrise} has been widely adopted in medical image analysis to learn transferable feature representations in the stage of pre-training. Recent state-of-the-art SSL approaches are often based on noise-contrastive estimation (NCE)~\cite{gutmann2010noise} (i.e., contrastive loss or contrastive learning), which learn transformation-invariant features by contrasting a few homogeneous image pairs with quite a large number (usually tens of thousands) of heterogeneous pairs \cite{zhou2021models,chen2020simple,he2020momentum} in each training step. However, the above characteristic may cause one intuitive problem in practice, which is the limited number of homogeneous pairs in contrastive analysis may prevent the model from capturing the most representative semantic features. 


On the other hand, in dermatology analysis, there are about 1,500 skin diseases and many variants that can be located in different body parts\footnote{\url{https://dermnetnz.org/cme/principles/an-overview-of-dermatology/}}. Although the medical imaging community has proposed several publicly available large-scale dermatology datasets, such as HAM10000 \cite{tschandl2018ham10000} and BCN20000 \cite{combalia2019bcn20000}, to address the limitation of data and labels for training neural networks, there still exists a great need of labeled dermatology images, especially when we consider the domain shift problem that often emerges in small-scale multi-domain, multi-center and multi-device datasets. From a different perspective, SSL provides a solution to utilize unlabeled dermatology data for pre-training, from which transferable images representations can be learned and serve as the feature basis for downstream small-scale datasets. 

In this paper, we introduce \textbf{Gr}ouping \textbf{a}ugmented \textbf{V}iews from \textbf{I}ndependent \textbf{S}ources (GraVIS) to learn self-supervised dermatology image representations. The core idea behind GraVIS is intuitive, which is good representations should be able to aggregate homogeneous image views while separating heterogeneous views. Here, the term \emph{homogeneous views} include one anchor view and at least one positive view, which are all generated by randomly augmenting the same source image. The \emph{heterogeneous views} involve one anchor view and at least one negative view, where each negative image view is generated by augmenting an image that is different from the one used for homogeneous views. Specifically, we use \emph{homogeneous pair} to refer to an image pair that consists of one anchor view and one positive view. Similarly, each \emph{heterogeneous pair} contains one anchor view and one negative view.

In Fig.~\ref{fig:intro}, we contrast our GraVIS with the canonical triplet loss~\cite{schultz2004learning} and NCE~\cite{gutmann2010noise}. In practice, GraVIS addresses the aforementioned major problem of NCE (i.e., few homogeneous pairs) by applying $N$-times augmentation to a given source dermatology image in order to generate one anchor view and $N$-1 positive views (i.e., $N$-1 homogeneous pairs). Meanwhile, GraVIS reduces the number of negative views to hundreds by taking into account one specific characteristic of dermatology images, which is heterogeneous dermatology image pairs are easy to recognize and thus we do not need a large number of heterogeneous pairs for contrast. Compared to the triplet loss, our method aims to aggregate multiple homogeneous views and separating heterogeneous views, simultaneously. Compared to NCE, GraVIS addresses more on grouping multiple homogeneous views for capturing the most general and transferable information hidden in the data. To achieve these goals, GraVIS introduce \textbf{V}iew \textbf{G}rouping \textbf{L}oss (VGL) to adaptively aggregate different types of views, where the hardness-aware attention is proposed to help the model focus more on homogeneous views with similar appearance by assigning larger penalty to them. To summarize, our contributions can be organized as follows:
\begin{itemize}
    \item A novel self-supervised learning methodology, GraVIS, is proposed for conduct unsupervised pre-training on dermatology images. GraVIS is inspired by the archetypical triplet loss and utilizes a similarity loss instead of the recent NCE to learn transferable dermatology image representations.
    \item In comparison to contrastive SSL approaches, GraVIS brings improvements by laying emphasis on homogeneous pairs instead of heterogeneous ones. We introduce a hardness-aware attention to adaptively assign different degrees of penalty to homogeneous pairs based on each pair's similarity score.
    \item Extensive experiments are conducted on two tasks: lesion segmentation and disease classification. Our GraVIS displays substantial and significant improvements over other SSL methods. More importantly, by fine-tuning the pre-trained weights acquired using GraVIS, we achieve quite competitive results using a \textsc{single} model on each task, even compared to the winners of ISIC 2017 challenges.
\end{itemize}

\section{Related work}
\subsubsection{Contrastive and non-contrastive self-supervised learning}
Contrastive learning aims to learn invariant representations of images via noise-contrastive estimation (NCE) \cite{gutmann2010noise}. Dosovitskiy \etal \cite{dosovitskiy2015discriminative} considered each image as its own class and used a linear classifier to classify each image. Wu \etal \cite{wu2018unsupervised} replaced the classifier with a memory bank that stores previously-computed representations as the former approach is unpractical in big datasets. Based on the memory bank, Misra \etal \cite{misra2020self} proposed to learn invariant representations based on pretext tasks. Similarly, Tian \etal \cite{tian2020contrastive} tried to maximize mutual information between different views of the same scene. CPC \cite{oord2018representation} and CPCv2 \cite{henaff2020data} combined predicting future observations (predictive coding) with a probabilistic contrastive loss. Chen \etal \cite{chen2020simple} introduced SimCLR, which shows that as long as the batch size for pre-training is large enough, well transferable image representations can be directly learned from the image batch without the memory bank. He \etal \cite{he2016deep} introduced MoCo, which improves the training of contrastive methods by comparing representations from the momentum network and the ordinary network. To replace the memory bank and solve the memory issue of SimCLR (caused by large batch sizes), MoCo utilizes a queue to store past representations. Caron \etal \cite{caron2020unsupervised} incorporated deep clustering strategies to facilitate contrastive learning by enforcing consistency between cluster assignments produced for different augmentations. Hjelm \etal \cite{hjelm2018learning} further showed the necessity of employing a large number of negative samples in the training batch to achieve satisfactory performance. Different from the above contrastive approaches that need a large number (tens of thousands) of heterogeneous pairs for contrast, our GraVIS is specifically designed for dermatology images, which lays more emphasis on homogeneous pairs and only requires hundreds of heterogeneous pairs to learn invariant and transferable image representations. On the other hand, there are some recent works \cite{grill2020bootstrap,chen2021exploring,zbontar2021barlow,li2021self,bardes2021vicreg} trying to replace the contrastive procedure with the siamese learning. Different from them, our GraVIS is inspired by the archetypical triplet loss.

\begin{figure}[t]
\centering
\includegraphics[width=0.9\columnwidth]{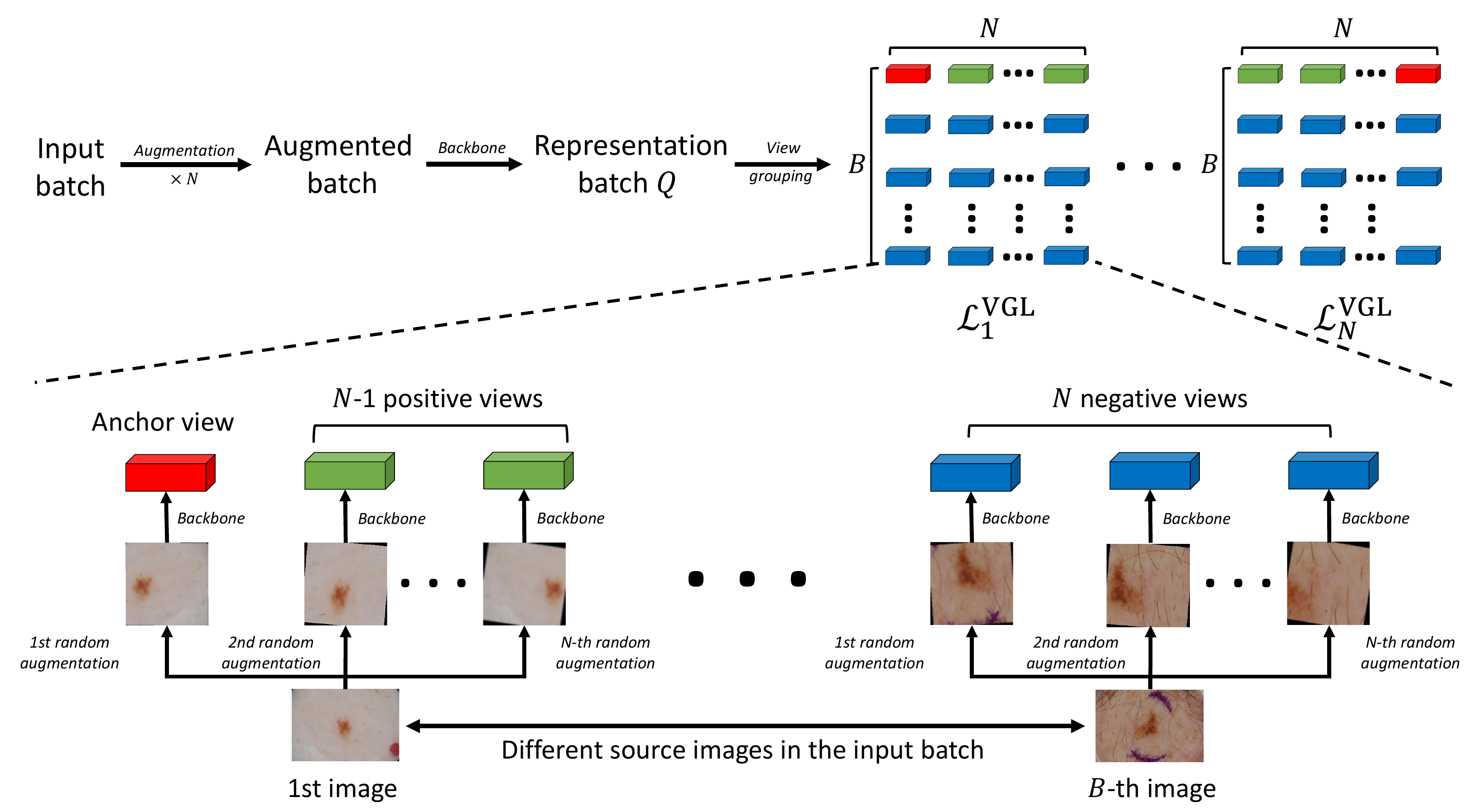} 
\caption{Overview of our proposed GraVIS. By augmenting the input image batch by $N$ times, we can acquire the augmented image batch, which is then passed to the backbone network to get a batch of representations, denoted as $Q$. Next, we apply view grouping loss to $Q$, where we randomly select a view representation as the anchor view (marked using red). We treat those representations that share the same source image with the anchor view as positive views (marked using green). Accordingly, representations that come from different source images are defined as negative views (marked using blue).} 
\label{Fig.framework}
\end{figure}

\subsubsection{Triplet loss and triplet mining}
The objective of the triplet loss is to minimize the distance between the anchor and positive sample while maximizing the distance between the anchor and negative sample. Schroff \etal \cite{schroff2015facenet} first applied the triplet loss to face recognition and verification using deep neural network. Cheng \etal \cite{cheng2016person} introduced an improved version of the triplet loss to pull the instances of the same person closer, and at the same time push the instances belonging to different persons farther from each other in the learned feature space. Xie \etal \cite{xie2020instance} used the triplet loss to learn representations for nuclei segmentation. Puch \etal \cite{puch2019few} implemented the triplet loss in the setting of few-shot learning to recognize brain imaging modality. When using the triplet loss to train neural networks, triplet mining is a widely adopted strategy to select hard \cite{hermans2017defense} or semi-hard \cite{schroff2015facenet,shi2016embedding} triplets for reducing the computational cost and accelerating training process. Oh Song \etal \cite{oh2016deep} incorporated batch-wise hard triplet mining into the training process of deep neural networks. In our GraVIS, we introduce an attention mechanism to help the network focus on the most similar homogeneous pairs while avoiding being dominated by dissimilar homogeneous pairs. This is different from the traditional triplet mining methods that often address the negative samples (i.e., heterogeneous pairs). It should be mentioned that the idea of addressing the importance of similar homogeneous pairs has been introduced by BagLoss~\cite{martinez2021training} to improve the performance of image retrieval models. The technical difference between BagLoss and GraVIS lies in the way to generate and utilize negative samples. BagLoss picks the hardest negative sample from the dataset, while GraVIS dynamically constructs negative samples using various augmentation strategies and uses all of them to learn invariant representations.

\subsubsection{Self-supervised learning in medical image analysis} Besides NCE-based SSL approaches, restoring/reconstructing the input images is the mostly adopted way to provide intrinsic supervision for learning self-supervised medical image representations. Chen \etal \cite{chen2019self} used context restoration to learn self-supervised representations on 2D ultrasound image, 3D abdominal Computerized Tomography (CT) scans and 3D brain Magnetic Resonance Imaging (MRI) scans. Zhou \etal \cite{zhou2021models} introduced Model Genesis that applies various augmentation strategies to build diverse reconstruction targets for both 2D and 3D medical images. Haghighi \etal \cite{haghighi2021transferable} modified Model Genesis by adding a classification branch to classify high-level features. Zhuang \etal \cite{zhuang2019self} introduced a jigsaw problem on reconstructing 3D brain MRI scans, which is then improved by \cite{zhu2020rubik} with a generative adversarial network to recover shuffled 3D patches. For NCE-based methods, Zhou \etal \cite{zhou2020comparing} proposed C2L which first applied NCE to 2D radiographs. C2L learns more invariant radiograph representations by contrasting homogeneous with heterogeneous pairs in both image- and feature-space. Taleb \etal \cite{NEURIPS2020_d2dc6368} investigated the performance of both NCE and self-reconstruction based approaches and found contrastive SSL methods are more advantageous. Besides self-supervised learning methodologies, Ke \etal \cite{ke2021chextransfer} made a range of experiments to study the influence of ImageNet-based pre-training on chest X-ray models. Different from the above works, we introduce a new SSL methodology which replies neither on NCE nor context restoration but surpasses both of them by substantial and significant margins. Recently, Azizi \etal \cite{azizi2021big} introduced a modified version of SimCLR (i.e., SimCLR-Derm) and a multi-instance version of SimCLR (i.e., MiCLE) to dermatology images. The second version utilizes the consistency of multiple images from a single patient. Vu \etal \cite{vu2021medaug} also employed a similar idea on X-ray data to make use of multiple views from the same patient to construct homogeneous pairs. For implementation, these two approaches require the access to the patient meta data in order to know which images come from the same patient. In contrast, our GraVIS does not have this constraint.

\section{Methodology}
\begin{figure}[t]
\centering
\includegraphics[width=0.75\columnwidth]{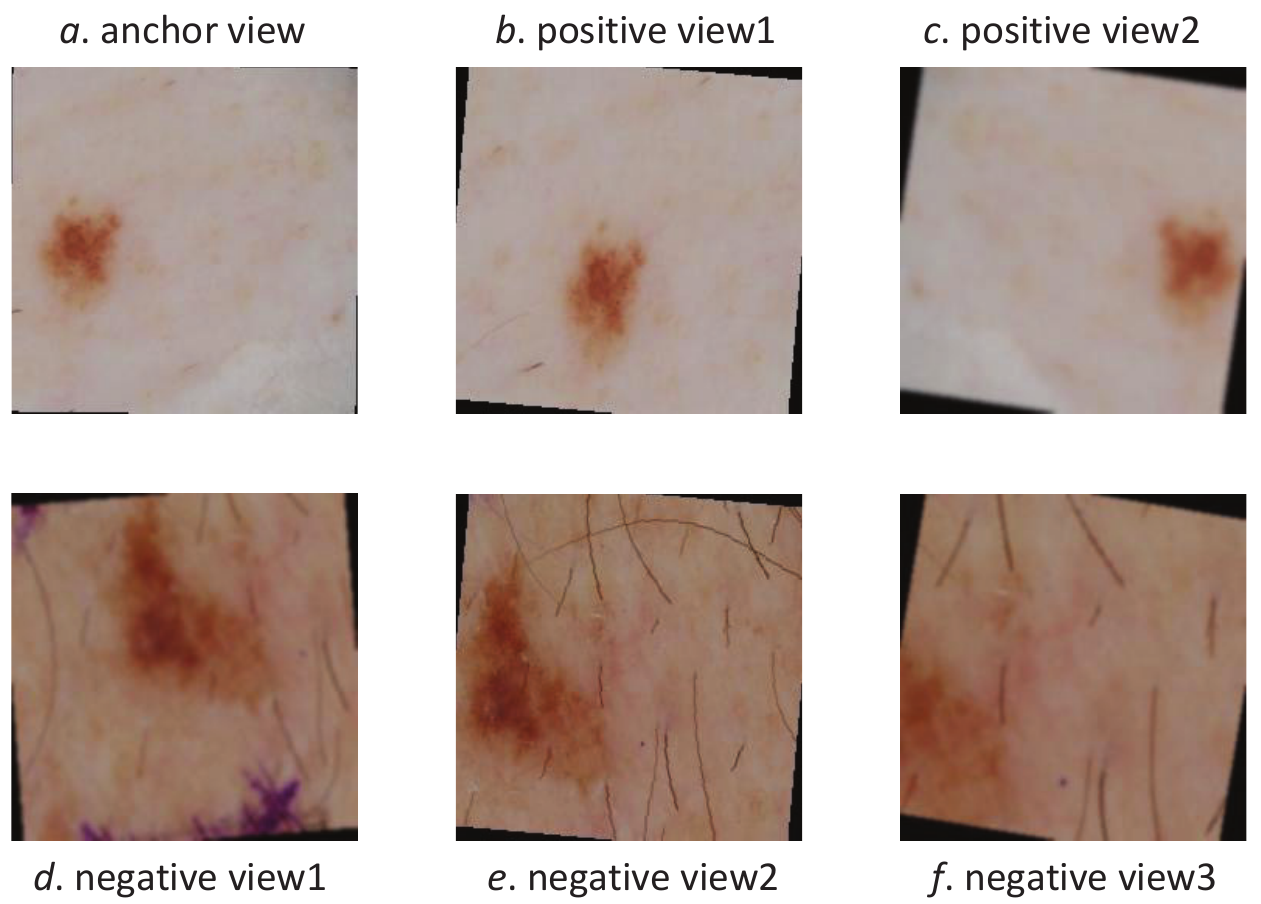}
\caption{An example of the formation of the anchor view, positive views and negative views. Note that views a, b and c come from the same source image while views d, e and f are from a different source image. Thus, view pairs (a, b) and (a, c) are homogeneous. Pairs (a, d), (a, e) and (a, f) are heterogeneous. All views are generated via different types and degrees of augmentation.}
\label{Fig.aug_ex}
\end{figure}

\subsection{Overview}
\textbf{Gr}ouping \textbf{a}ugmented \textbf{V}iews from \textbf{I}ndependent \textbf{S}ources (GraVIS) aims at conducting pre-training using a large number of unlabeled dermatology images, after which we fine-tune the pre-trained model on different downstream tasks to verify the transferable ability of learned representations. We provide an overview of our GraVIS  in Fig. \ref{Fig.framework}. In each training step, we first forward a batch of dermatology images to the backbone network in order to acquire their representations, on top of which the view grouping loss (VGL) is applied to aggregate homogeneous samples while separating heterogeneous ones. Compared to recent contrastive learning based approaches \cite{azizi2021big,zhou2020comparing}, GraVIS does not need any contrastive operations but still displays state-of-the-art performance on a variety of downstream tasks.

\subsection{Positive and negative sets}
Given an input image batch, we first randomly augment each image in this batch by $N$ times. The augmentation strategies mainly include random crop, random horizontal flip, random rotation and color distortion. We provide an example in Fig. \ref{Fig.aug_ex} where we display one anchor view, two positive views and three negative views. We apply random crop and random rotation (a small degree) to the source dermatology image to acquire the anchor view (\emph{a}). To generate the positive view1 (\emph{b}), we include random crop and random rotation with a relatively large degree. For positive view2 (\emph{c}), we include random crop, horizontal flip, random rotation and gaussian blur. Similar combinations of augmentation strategies can also be observed in negative views (\emph{d}, \emph{e} and \emph{f}).
Supposing the size of the input image batch is $B$ (we ignore the height, width and channel number of images for simplicity), after $N$-times random augmentation, we obtain an enlarged image batch whose size is $B$*$N$. During the pre-training stage, in each training step, we pass this enlarged batch to the backbone network to acquire a batch of corresponding visual representations (denoted as batch $Q$ in the following), from which we randomly pick a representation and regard it as the anchor representation. Given this anchor view, its positive set contains the rest $N-1$ views (i.e., representations) that are derived from the same source image. Intuitively, the rest $(B-1)$*$N$ views construct the negative set as they correspond to different source dermatology images. Based on positive and negative sets, we calculate VGL to minimize the distance between the anchor view and its positive set while keeping the anchor away from those views in the negative set. 

\subsection{View grouping loss}
\begin{figure}[t]
    \centering
    \includegraphics[width=0.75\columnwidth]{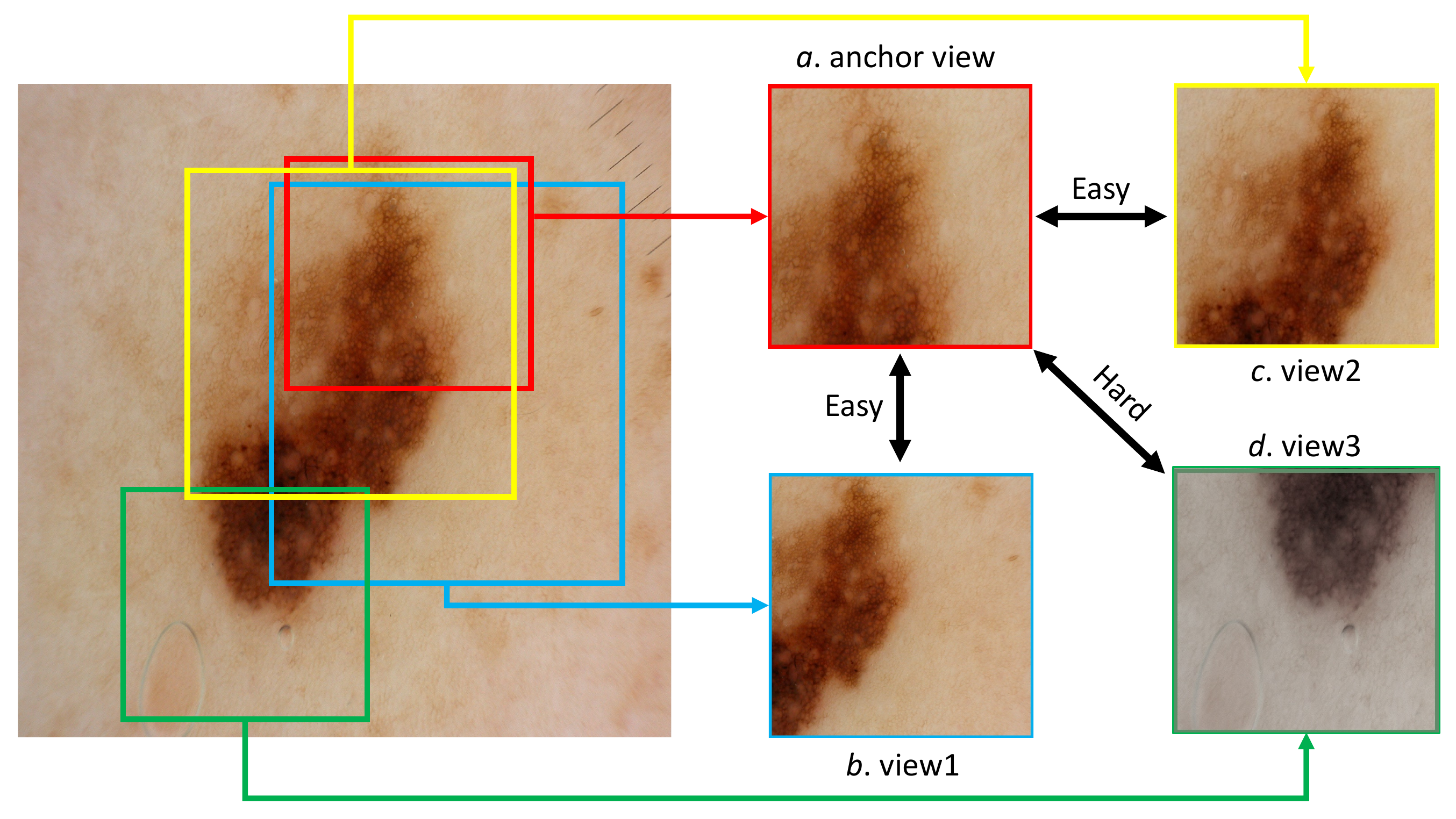}
    \caption{Hardness of recognizing homogeneous image pairs based on appearance. It is observable that the anchor view (\emph{a}), view1 (\emph{b}) and view2 (\emph{c}) are from the same source image as they look similar in appearance. However, although view3 (\emph{d}) is randomly cropped from the same source image as that of the anchor view, we can hardly tell whether they come from the same source without expertise as view3 has quite different appearance and there is no clear overlap between them. Thus, it may not be reasonable to directly ask the representation of view3 to be as close to the representation of anchor view as other positive features.}
    \label{loss_motivation}
\end{figure}
In GraVIS, we define the view grouping loss (VGL) to be:
\begin{align}
\begin{split}
    \label{Equation.1}
    \mathcal{L}^{\text{VGL}}_q & = 1 - \frac{1}{\|\mathcal{S}^{\text{ps}}\|} \smashoperator{\sum_{i\in \mathcal{S}^{\text{ps}}}} \frac{1}{\gamma_{q,i}\smashoperator{\sum_{j \in \mathcal{S}^{\text{ns}}}}\sigma (c_{q, j} - c_{q, i})+1},\\
    \gamma_{q,i} & = \frac{1}{\smashoperator{\sum_{j \in \mathcal{S}^{\text{ps}}, j \neq i}} \sigma (c_{q, j} - c_{q, i})}.
\end{split}
\end{align}
$\mathcal{L}^{\text{VGL}}_q$ stands for VGL with $q$-th sample in batch $Q$ as the anchor view. Accordingly, $\mathcal{S}^{\text{ps}}$ denotes the positive set and $\mathcal{S}^{\text{ns}}$ stands for the negative set. $\|\mathcal{S}^{\text{ps}}\|$ refers to the number of views in $\mathcal{S}^{\text{ps}}$. $c_{i,j}$ measures the similarity between the $i$-th sample and the $j$-th one, which is calculated using the cosine similarity:
\begin{equation}
\label{equation.cosine}
    c_{i,j} = \frac{\textbf{v}_i^T \textbf{v}_j}{\Vert{\textbf{v}_i}\Vert \Vert{\textbf{v}_j}\Vert},
\end{equation}
where $\textbf{v}_i$ and $\textbf{v}_j$ correspond to the $i$-th and $j$-th feature vectors in batch $Q$, respectively. In Equation \ref{Equation.1}, $\sigma(\cdot)$ stands for the sigmoid function:
\begin{equation}
\label{equation.sigmoid}
    \sigma (x) = \frac{1}{1 + e^{\frac{-x}{\tau}}}
\end{equation}
with the temperature hyper-parameter $\tau$ controlling how much GraVIS addresses similar image pairs.

\begin{algorithm}[t] 
	\renewcommand{\algorithmicrequire}{\textbf{Input:}}
	\renewcommand{\algorithmicensure}{\textbf{Output:}}
	\caption{GraVIS:}
	\label{alg1}
	\begin{algorithmic}[1]
		\STATE \textbf{Notations}: \textsc{Net} denotes the backbone network. $K$ stands for the number of training steps. \textsc{iter} is the index of iteration.
		\FOR{$\textsc{iter}\leftarrow 0$ to $K$}
		\STATE Randomly sample a batch of images $\mathcal{X}_o$;
		\STATE Augment $\mathcal{X}_o$ by $N$ times to acquire an enlarged image batch $\mathcal{X}_e$;
		\STATE Pass $\mathcal{X}_e$ to \textsc{Net} to get $Q$;
		\STATE Initialize $\mathcal{L}_Q$ to 0;
		\FOR{$q \leftarrow 0 $ to $ B * N$}
		\STATE $\mathcal{L}_Q$ += $\mathcal{L}^{\text{VGL}}_q.$;
		\ENDFOR
		\STATE Backward ($\mathcal{L}_Q$) to update \textsc{Net};
        \ENDFOR
	\end{algorithmic}  
\end{algorithm}
\begin{figure}[t]
\centering
\subfigure[Impact of $\gamma$ on the gradient]{
\label{gamma_grad}
\includegraphics[width=0.48\columnwidth]{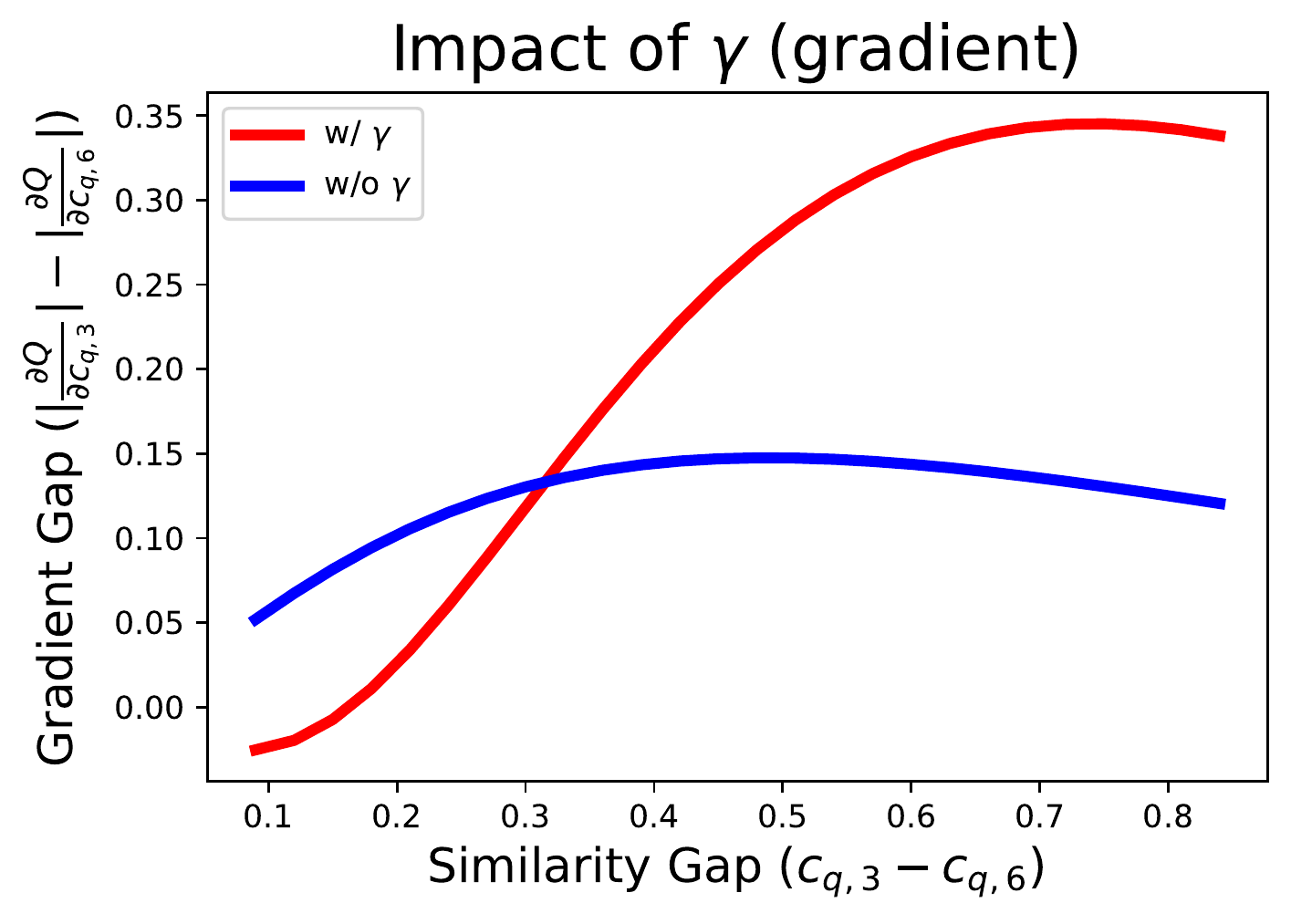}}
\subfigure[Impact of $\gamma$ on the loss value]{
\label{gamma_loss}
\includegraphics[width=0.46\columnwidth]{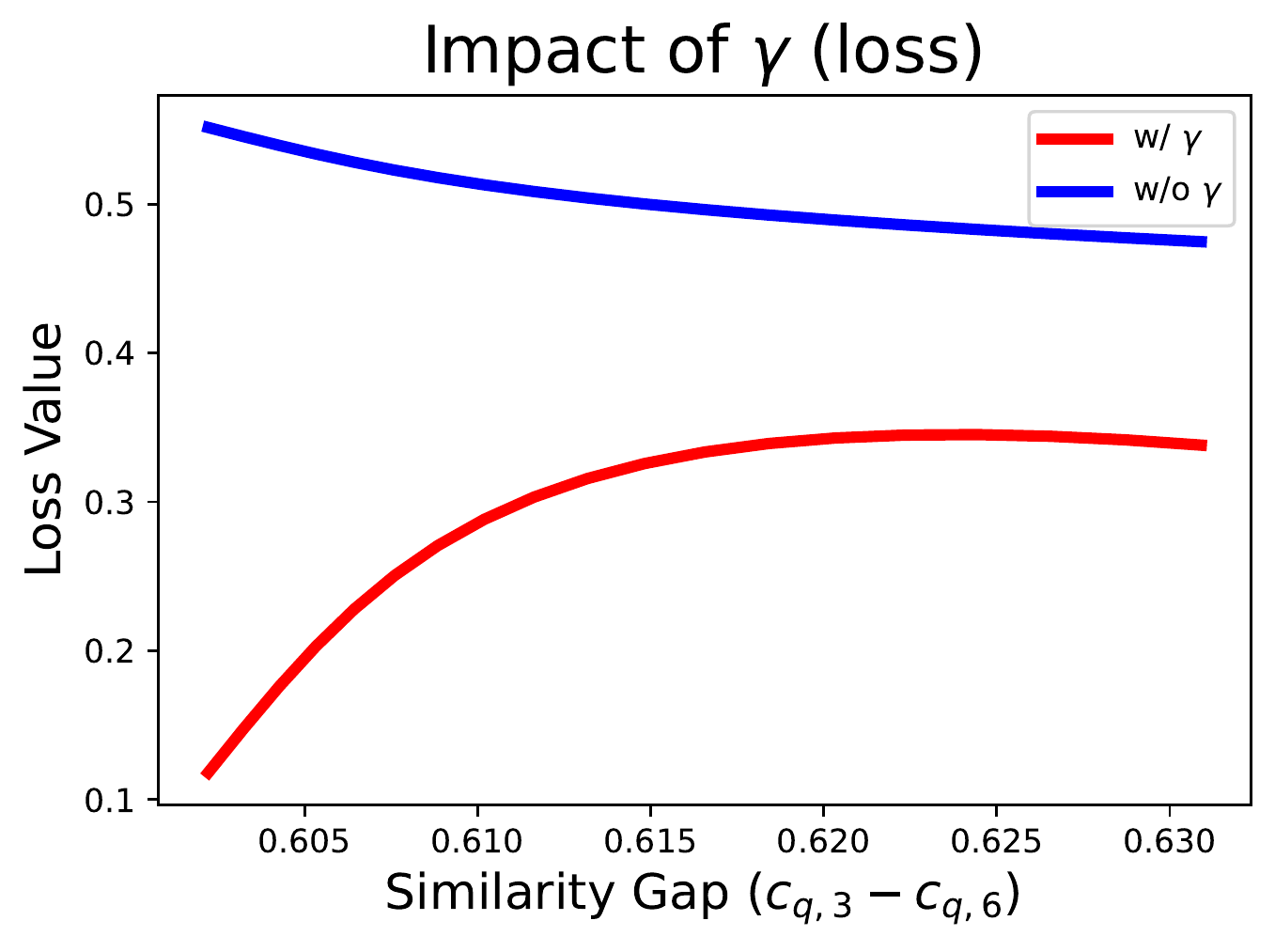}}
\caption{Impact of the hardness-aware attention $\gamma$. The left figure (a) shows that the introduced attention $\gamma$ focuses more on the similar homogeneous pair ($\textbf{v}_q$, $\textbf{v}_3^{ps}$) (the similarity score is $c_{q,3}$) instead of the dissimilar pair ($\textbf{v}_q$, $\textbf{v}_6^{ps}$) (the similarity score is $c_{q,6}$). The right figure (b) demonstrates that adding $\gamma$ makes the view grouping loss more sensitive to the similarity gap between two homogenenous pairs.}
\label{imp_gamma}
\end{figure}

Term $\smashoperator{\sum_{j \in \mathcal{S}^{\text{ns}}}} \sigma (c_{q, j} - c_{q, i})$ evaluates $c_{q,i}$ among the negative set ($\mathcal{S}^{\text{ns}}$). By aggregating the anchor view ($q$) and positive view ($i$) and separating the anchor view from negative views ($\forall j\in \mathcal{S}^{\text{ns}}$), $\smashoperator{\sum_{j \in \mathcal{S}^{\text{ns}}}} \sigma (c_{q, j} - c_{q, i})$ becomes smaller and $\mathcal{L}^{\text{VGL}}_q$ is accordingly minimized.

The hardness-aware attention $\gamma_{q,i}$ evaluates $c_{q,i}$ among the positive set ($\mathcal{S}^{\text{ps}}$). The intuition behind is that we believe more attention should be allocated to homogeneous pairs of similar views (in appearance) instead of dissimilar (though homogeneous) pairs. We give an example to illustrate our intuition in Fig. \ref{loss_motivation}, where one anchor view and three positive views are displayed. It is easy to recognize the relationship between the anchor view and view1 (\emph{b}) as both of them are overlapped after being cropped from the same source image. Similar phenomenon can also be observed between the anchor view and view2 (\emph{c}). However, when we compare the anchor view with view3 (\emph{d}), we can hardly tell whether they two share the same origin without relying on expertise because they have quite dissimilar appearance. In fact, it may not be suitable to require the representation of view3 to be as close to the anchor representation as other homogeneous pairs (i.e., forcing $c_{a,3} \approx c_{a,1} \approx c_{a,2}$ is questionable). To address this issue, the hardness-aware attention $\gamma_{q,i}$ tends to assign a large penalty to similar homogeneous pairs by taking into account their similarity ($c_{q,i}$) in all homogeneous pairs ($c_{q,j}$, $\forall j \in \mathcal{S}^{\text{ps}}, j \neq i$). In practice, the hardness-aware mechanism helps GraVIS to focus more on recognizing potentially ``easy'' relationships instead of regularizing meaningless dissimilar homogeneous pairs. 

We provide an example to illustrate the impact of $\gamma$ on VGL. Considering three positive homogeneous view vectors (i.e., $S^{\text{ps}}$=$\{\textbf{v}^{\textbf{ps}}_1,\textbf{v}^{\textbf{ps}}_3,\textbf{v}^{\textbf{ps}}_6\}$) and three negative view vectors (i.e., $S^{\text{ns}}$=$\{\textbf{v}^{\textbf{ns}}_2,\textbf{v}^{\textbf{ns}}_4,\textbf{v}^{\textbf{ns}}_5\}$), all of which are ordered based on the similar score with the anchor view in a descending order, i.e.,
\begin{align}
    \begin{split}
        &\{\textbf{v}^{ps}_1,\ \textbf{v}^{ns}_2,\ \textbf{v}^{ps}_3,\ \textbf{v}^{ns}_4,\ \textbf{v}^{ns}_5,\ \textbf{v}^{ps}_6\},\\
        s.t.\ &\forall i \in \{1,2,3,4,5\}, c_{q,i} - c_{q,i+1} = C,
    \end{split}
    \label{example}
\end{align}
where $c_{q,i}$ stands for the similarity score between the $i$-th view and the anchor view. For simplicity, we define the score interval between two adjacent views to be a positive constant $C$. As aforementioned, the proposed hardness-aware attention $\gamma$ helps VGL to lay more emphasis on shifting easy samples. In Fig.~\ref{gamma_grad}, we can see that as the similarity gap between $c_{q,3}$ and $c_{q,6}$ becomes larger, VGL (with the attention factor $\gamma$) assigns larger gradients (denoted as $|\frac{\partial Q}{\partial c_{q,3}}|-|\frac{\partial Q}{\partial c_{q,6}}|$) to the more similar homogeneous pair ($\textbf{v}_q$, $\textbf{v}_3^{ps}$) instead of the less similar homogeneous pair ($\textbf{v}_q$, $\textbf{v}_6^{ps}$). Moreover, in Fig.~\ref{gamma_loss}, we observe that equipped with $\gamma$, VGL becomes more sensitive to the similarity gap between two homogenenous pairs, which again verifies the impact of the hardness-aware attention on handling dissimilar homogeneous pairs. 


Finally, for each batch $Q$, we calculate the overall loss by summing up all $\mathcal{L}^{\text{VGL}}_q$, which can be presented as follows:
\begin{align}
    \begin{split}
        \mathcal{L}_Q = \sum_{q=1}^{B*N} \mathcal{L}^{\text{VGL}}_q.
    \end{split}
    \label{loss_Q}
\end{align}

\noindent \subsubsection{Comparison with noise-contrastive estimation} Here we would like to clarify the advantages of proposed VGL over the noise-contrastive estimation (NCE) \cite{gutmann2010noise}, which has been widely used in self-supervised visual representation learning \cite{chen2020simple,zhou2020comparing,he2020momentum}. Similar to VGL, NCE contrasts the homogeneous pair with multiple heterogeneous pairs to learn invariant representations; however, it suffers from two issues when dealing with dermatology images: i) the locality problem and ii) the equality problem. For issue i), given an anchor view in the input image batch, the archetypical NCE compares it with only one homogeneous view (i.e., one homogeneous pair), which cannot guarantee the learned semantic information are robustness enough towards a variety of transformations (i.e., augmentation strategies). For issue ii), NCE treats all homogeneous pairs equally regardless of the degree of applied augmentations. As we have mentioned in Fig. \ref{loss_motivation}, for dissimilar homogeneous views (e.g., the anchor view and view3), it may not be appropriate to force their representations to be close to each other. In contrast, our GraVIS addresses issue i) by introducing more homogeneous pairs to learn more invariant semantic representations. To tackle issue ii), VGL comprises the hardness-aware attention to assign adaptive penalty on top of the hardness of recognizing the relationship between two homogeneous views.\\

\subsubsection{Comparison with the triplet loss} Each time, the triplet loss contrasts one homogeneous pair with one heterogeneous pair as follows:
\begin{equation}
     L_{triplet}(\textbf{v}_q, \textbf{v}_{ps}, \textbf{v}_{ns}) = max(0, m + d(\textbf{v}_q, \textbf{v}_{ps}) - d(\textbf{v}_q, \textbf{v}_{ns}))
     \label{equation.4}
\end{equation}
where $\textbf{v}_q$, $\textbf{v}_{ps}$ and $\textbf{v}_{ns}$ denote the query representation (i.e., the anchor view), positive and negative representations, respectively. $m$ is a pre-defined margin. $d$ stands for the distance function, where euclidean distance is mostly used and cosine similarity is also applicable. Compared to GraVIS, the triplet loss does not require to apply $N$-times augmentation, which lead to only one homogeneous pair and one heterogeneous pair in each computation. Thus, the learned representations are much less invariant than those learned with VGL or NCE, which may be the reason why the triplet loss has not been widely used in self-supervised representation learning. Another drawback of the triplet loss is that it linearly combines each term, ignoring the difference and importance of different homogeneous or heterogeneous pairs. As a result, the training process may be less effective as it is dominated by the trivial information from a few meaningless yet hard samples. For instance, it is possible that $\textbf{v}_q$ and $\textbf{v}_{ps}$ originate from two augmented images that share the same source image but have dramatically different appearance (such as view1 and view3 in Fig. \ref{loss_motivation}). In this case, minimizing the distance between $\textbf{v}_q$ and $\textbf{v}_{ps}$ becomes meaningless, especially when $\textbf{v}_{ns}$ comes from a dermatology image that suffers from the same skin disease as that of $\textbf{v}_q$ and $\textbf{v}_{ps}$. In contrast, our GraVIS is able to reduce the importance of such dissimilar homogeneous pairs using the hardness-aware attention.

\section{Experiment}
\begin{table*}[]
    \renewcommand{\arraystretch}{1.3}
    \caption{Experiments in lesion segmentation and disease classification tasks. Different ratios stand for different amounts of data for fine-tuning. We report the jaccard index for lesion segmentation and mean accuracy for disease classification.}
    \setlength{\abovecaptionskip}{10pt}%
    \centering
        \begin{tabular}{c|c|c|c|c|c|c|c|c|c|c|c|c}
            \toprule
            &\multicolumn{6}{c|}{Lesion segmentation} & \multicolumn{6}{c}{Disease classification}\\
            \hline
            Methodology &10\%&20\%&30\%&40\%&50\%&100\%&10\%&20\%&30\%&40\%&50\%&100\%  \\
            \hline
            \hline
            TS & 58.9&61.3&63.4&69.5&70.5 & 72.3&78.3&79.7&80.4&81.8&82.5&85.9\\
            \hline
            IN&68.4&70.1&71.8&72.1&72.5&74.1&80.1&83.5&84.1&84.3&85.1&87.0\\
            \hline
            NPTL&66.7&68.9&70.4&70.8&71.1&73.9&79.2&81.7&83.4&83.8&84.2&86.6\\
            \hline
            SimCLR-Derm & 67.7&69.6&70.8&71.4&71.5& 73.9& 80.5&83.8&84.3&84.6&85.3&86.5\\
            \hline
            MoCov2 & 68.0 & 70.0 & 71.9 & 72.3 & 72.6 & 74.5 & 83.1 & 83.8 & 85.0 & 85.7 & 85.8 & 87.8 \\
            \hline
            BYOL & 68.3 & 69.6 & 72.2 & 72.7 & 73.0 & 74.9 & 81.4 & 83.3 & 85.3 & 85.8 & 86.0 & 86.7 \\
            \hline
            BagLoss & 63.7 & 68.8 & 69.5 & 70.8 & 71.2 & 73.3 & 76.8 & 80.3 & 81.2 & 81.7 & 82.4 & 83.5 \\
            \hline
            C2L&68.9&70.7&71.9&72.4&72.5&74.4&81.3&84.7&85.2&85.7&85.8&87.6 \\
            \hline
            MG&66.2&68.9&70.3&70.7&71.3&72.8&77.1&78.7&79.2&80.8&81.7&85.9 \\
            \hline
            Our GraVIS &\textbf{70.9}&\textbf{72.6}&\textbf{73.0}&\textbf{73.3}&\textbf{74.0} & \textbf{75.7} & \textbf{85.9}&\textbf{86.4}&\textbf{86.7}&\textbf{86.9}&\textbf{87.0}&\textbf{89.1}\\
            \bottomrule
    \end{tabular}
    \label{tab:class}
\end{table*}

\subsection{Dataset} 
We evaluate GraVIS on ISIC 2017 dataset \cite{codella2018skin}. The numbers of images in the training, validation and test sets are 2000, 150 and 600, respectively. Lesions in dermatology images are all paired with a gold standard (definitive) diagnosis, i.e. melanoma, nevus, and seborrheic keratosis. There are three major tasks in ISIC 2017 Challenge and we evaluate our framework on lesion segmentation (task1) and disease classification (task2). In disease classification, we conduct two binary classification sub-tasks: i) distinguishing between (a) melanoma and (b) nevus and seborrheic keratosis; ii) distinguishing (a) seborrheic keratosis from (b) nevus and melanoma. The evaluation metrics are jaccard index and mean accuracy for task1 and task2, respectively, which are also used in the released official leaderboard of ISIC 2017.
\subsection{Implementation details}
\subsubsection{Details of pre-training} We first remove the labels of all dermatology images in the training set to pre-train the backbone network using GraVIS. Specifically, the size of the input image batch is 32 and the number of random augmentation $N$ is 20. We employ SGD with momentum as the default optimizer whose initial learning rate is set to 1e-3 while the momentum value is set to 0.9. We employ the cosine annealing strategy for learning rate decay and train the network for 240 epochs. We use ResNet50 \cite{he2016deep} with two fully-connected layers as the backbone network whose output dimension is 1$\times$128. The pre-training procedure is finished on 4 NIVIDA TITAN V GPUs within 8 hours. In each training step, the augmented image batch contain $B$*$N$ images, where all positive images are continuous and thus the model is likely to find a short cut by directly making use of the location information (i.e., the model knows which images are from the same source and which are not). So we shuffle the batch before we pass it to the backbone network, which prevents the network from finding a trivial solution that may lead to trivial representations.

\subsubsection{Details of fine-tuning} In the fine-tuning stage, we remove two fully-connected layers used in the pre-training stage and load the weight parameters of convolutional layers. In task1, we use ResU-Net \cite{zhang2018road} for lesion segmentation, where we load pre-trained ResNet as the encoder and randomly initialize the corresponding decoder. The batch size for segmentation task is 32. The loss function includes both dice loss and cross entropy loss whose coefficients are 1 and 0.2, respectively. In task2, we add a linear classification head with a dropout ($p$ = 0.2) layer before it. The batch size of classification task is 64 and the loss function is binary cross entropy loss. For all tasks, we unfreeze the weights of convolutional layers and fine-tune the entire network. Validation score (accuracy or jaccard index) is obtained after each training epoch and then we anneal the learning rate by a factor of 0.5 if the validation score is not improved after 3 epochs. Different from the pre-training stage, we employ Adam as the optimizer and the initial learning rate is set to 1e-4. We stop the fine-tuning process when the validation score does not decrease for 10 epochs, and we save the checkpoint with the lowest validation loss value for testing. To evaluate the ability of GraVIS under different amounts of labeled data, we randomly sample 10\% to 50\% labeled data from the whole training set to fine-tune the model.

\subsection{Baselines}
Training from scratch (TS) and ImageNet-based initialization (IN) are involved as two basic baselines. For contrastive approaches, we include SimCLR-Derm~\cite{azizi2021big}, MoCov2~\cite{chen2020improved}, BYOL~\cite{grill2020bootstrap} and C2L~\cite{zhou2020comparing} for comparison. As for traditional self-supervised learning methods, we compare GraVIS against Model Genesis (MG)~\cite{zhou2021models}, $N$-pairs triplet loss with the triplet mining strategy (NPTL)~\cite{sohn2016improved} and BagLoss~\cite{martinez2021training}, where the last two methods are based on metric learning.

\subsection{Boosting performance under limited supervision}
\subsubsection{Lesion segmentation} We present experimental results of fine-tuning under different labeling ratios in Table \ref{tab:class}. Firstly, it is obvious that all representation learning (i.e., transfer learning and self-supervised learning) methods can distinctly boost the performance compared to TS, verifying the necessity of learning transferable representations during the pre-training stage. Secondly, we can easily find that self-supervised approaches based on NCE are advantageous over the predictive method (i.e., MG) under different ratios, which implies that NCE does perform better than the traditional predictive representation learning for dermatology images. Somewhat surprisingly, we found that an improved multi-pair version of triplet loss (i.e., NPTL) can already achieve comparable results with MG. Considering MG needs various pre-defined restoration tasks, such phenomenon suggests the potential of applying the triplet loss to self-supervised representation learning. Among all NCE-based baselines, BYOL achieves the best performance in large labeling ratios while C2L performs better in small ratios. When comparing our GraVIS against other baselines, it is observable that GraVIS has the ability to outperform other baselines in various ratios by obvious margins. More importantly, GraVIS surpasses the best performing self-supervised baselines C2L and BYOL by 2 percents under extremely limited supervision (10\% and 20\%) without explicitly using any contrastive functions. Compared to IN using 100\% labeled data (the most widely used method for tackling limited annotations), GraVIS achieves comparable results using only 50\% labeled dermatology images. It is worth noting that similar to the introduced hardness-aware attention, BagLoss addresses the importance of aggregating multiple homogeneous pairs in image retrieval, which is the reason why we include it as one of our baselines. From Table~\ref{tab:class}, we can see that the performance of BagLoss are not satisfactory. The reason behind is that BagLoss only assigns one hard negative sample to each anchor view, which may prevent the model from learning invariant and discriminative representations from a number of negative views.\\

\noindent \textbf{Statistical significance.} A t-test validation is conducted in all labeling ratios. The p-values between GraVIS and C2L/BYOL are smaller than 0.01, which indicates that GraVIS is statistically better than C2L/BYOL at the 1\% significance level. In addition, the p-values between GraVIS (using 50\% annotations) and IN (using 100\% labeled data) are much larger than 0.05, which shows that GraVIS produces competitive results compared to the most widely used fully-supervised baseline (IN-100\%).\\

\subsubsection{Disease classification} The best performing baselines in this task are MoCov2 and C2L, where MoCov2 maintains relatively obvious advantages when the labeling ratio is 10\%. In comparison to the performance on lesion segmentation, our GraVIS maintains relatively larger advantages over different baselines in the problem disease classification. For instance, GraVIS outperforms MoCov2 by nearly 3 percents when using 10\% labeled data. We believe these large improvements brought by GraVIS can be attributed to the strong demand of annotations in distinguishing three diseases (i.e., melanoma, nevus and seborrheic keratosis), where GraVIS can make use of limited annotations more effectively. Compared to the most widely adopted IN-100\%, our approach again produces quite comparable performance by using only half labeled data. These phenomena suggest the potential of GraVIS to replace canonical self-supervised and transfer learning based methods to learn more effective pre-trained representations for dermatology images.\\

\noindent \textbf{Statistical significance.} The p-values between GraVIS and the best performing baselines C2L/MoCov2 are smaller than 0.01 under different labeling ratios. Similar to the results of lesion segmentation, we believe these p-values help verify the statistical effectiveness of GraVIS over C2L/MoCov2. Again, the p-values between GraVIS-50\% and IN-100\% are much larger than 0.05, suggesting that GraVIS can greatly reduce the amount of human annotations that are necessary for training a fully-supervised diagnosis model that is based on ImageNet pre-training.

\subsection{Advancing fully-supervised fine-tuning}
To demonstrate that our GraVIS can boost the performance of fully-supervised fine-tuning (denoted as GraVIS (100\%) in Table~\ref{tab:top5}), we present a comparison of our method with top-5 results in the official leaderboard\footnote{https://challenge.isic-archive.com/leaderboards/2017} and recent state-of-the-art approaches~\cite{zhang2019attention,xie2020mutual}. Like~\cite{zhang2019attention,xie2020mutual}, we also collected 1320 additional dermoscopy images, including 466 melanoma, 822 nevus images, and 32 seborrheic keratosis images, from the ISIC Archive\footnote{https://www.isic-archive.com/} to enlarge the training dataset. We also incorporate advanced pre- and post-processing (PP) techniques to improve the model performance. For both tasks, we apply aggressive rotation augmentation to each image, where we rotate each image 15 degrees more on the basis of the previous one. Thus, we have 24 (i.e., $\frac{360}{15}$) rotated versions given a source image. Note that the accompanying segmentation mask is also rotated and aligned with the rotated input. Specifically, for lesion segmentation, we add L (lightness) channel in CIELAB space \cite{hunter1948accuracy} for better capturing the lesion boundary. We then resize each image to 192$\times$256 (height$\times$width), where a similar aspect ratio (3:4) is mostly used in the training set. To produce consistent predictions, we first use a high-confidence threshold (0.8) to determine the center of lesion, and then a middle-threshold (0.5) is used to highlight the lesion area. After filling holes, we ask the lesion area to contain the detected lesion center and take the maximum connected area as the final prediction. In practice, our GraVIS firstly pre-trains the ResU-Net (segmentation)/ResNet50 (classification) using the whole ISIC 2017 training set and the external data (without labels). Then, we fine-tune the pre-trained models on the labeled training set and external data.
\begin{table}[t]
    \footnotesize
    \centering
    \caption{Comparison with top-5 approaches in the leaderboard of ISIC 2017 and recent state-of-the-art methods. Note that we only use single models for testing on both tasks. 50\% and 100\% stand for the labeling ratios of annotated data used in the fine-tuning stage.}
    \renewcommand{\arraystretch}{1.3}
    \setlength{\abovecaptionskip}{10pt}%
    \begin{tabular}{c|c|c|c|c}
        \toprule
        Rank & \multicolumn{2}{c|}{Lesion segmentation}& \multicolumn{2}{c}{Disease classification}  \\
        \hline
        1 & Mt.Sinai~\cite{yuan2017automatic}&76.5&Casio and Shinshu~\cite{matsunaga2017image}& 91.1\\
        \hline
        2 & NLP LOGIX~\cite{berseth2017isic} &76.2&Multimedia~\cite{diaz2017incorporating}&91.0\\
        \hline
        3 & USYD-BMIT~\cite{bi2017automatic} &76.0&RECOD Titans~\cite{menegola2017recod}&90.8\\
        \hline
        4 & USYD-BMIT~\cite{bi2017automatic}&75.8&USYD-BMIT~\cite{bi2017automatic}&89.6\\
        \hline
        5 & RECOD Titans~\cite{menegola2017recod} &75.4& A*STAR~\cite{yang2017novel}&88.6\\
        \hline
        & Xie \etal~\cite{xie2020mutual}& 80.4 & Bisa \etal~\cite{bisla2019towards}&91.5\\
        \hline
        &GraVIS (50\%)&80.2 & Zhang \etal~\cite{zhang2019attention}&91.7\\
        \hline
        &GraVIS (100\%)&\textbf{81.0} & GraVIS (50\%) & 91.4\\
        \hline
        & & & GraVIS (100\%) & \textbf{92.1}\\
        \bottomrule
    \end{tabular}
    \label{tab:top5}
\end{table}
As shown in Table \ref{tab:top5}, GraVIS achieves the highest jaccard index in lesion segmentation and ranks 1st in the disease classification task, outperforming approaches in the leaderboard and recent state-of-the-art methods. We believe these observations show that GraVIS has the potential for advancing the performance of fully-supervised models.

To demonstrate the effectiveness of GraVIS in reducing the amount of labeled data, we also present the results of fine-tuning GraVIS-based pre-trained models with 50\% annotated data in Table~\ref{tab:top5} (denoted as GraVIS (50\%)). We see that GraVIS trained with only 50\% annotated data performs comparably with the fully-supervised state-of-arts in both lesion segmentation (i.e., \cite{xie2020mutual}) and disease classification (i.e., \cite{zhang2019attention}) tasks. We believe these results show the potential of GraVIS for serving as an annotation-efficient learning methodology.

\subsection{Ablation study}
In this section, we conduct ablative experiments for different modules and hyper-parameters to investigate their significance towards the performance of GraVIS. Specifically, the study targets consist of the hardness-aware attention, two factors that affect the number of positive and negative views: the number of augmentation $N$ and the input batch size $B$. We also display the influence of the slope controller $\tau$ in Eq. \ref{equation.sigmoid}. Last but not least, we demonstrate the effects of pre-training with respect to the number of pre-training epochs. Note that all ablation studies are fine-tuned with 100\% labeled data in the task of lesion segmentation.

\subsubsection{Influence of hardness-aware attention $\gamma$} In Table \ref{tab:att}, we report the results equipped with $\gamma$ and without $\gamma$. It is apparent that the proposed hardness-aware attention can bring substantial improvements in the setting of fully-supervised fine-tuning. We believe the underlying reason is that $\gamma$ prevents the model from laying too much emphasis on hard homogeneous pairs with quite different appearance that would cover up the effects of similar homogeneous pairs and dominate the training process. Without $\gamma$, GraVIS performs slightly worse than SimCLR-Derm but still better than MG, which indicates that GraVIS could become a useful self-supervised learning method even without the proposed attention mechanism.
\begin{table}[]
    \centering
    \caption{Influence of introduced hardness-aware attention $\gamma$. \textbf{w/} and \textbf{w/o} denotes with and without, respectively.}
    \renewcommand{\arraystretch}{1.3}
    \setlength{\abovecaptionskip}{10pt}%
    \begin{tabular}{c|c|c}
    \toprule
    Method &w/ $\gamma$ & w/o $\gamma$ \\
    \hline
    Jaccard index (\%)&\textbf{75.7} & 73.4 \\
    \bottomrule
    \end{tabular}
    \label{tab:att}
\end{table}

\subsubsection{Investigation of N-times augmentation} One of the core ideas of GraVIS is to contrast $N-1$ homogeneous pairs (excluding the anchor view) with $(B-1)$*$N$ heterogeneous image pairs. Intuitively, a larger $N$ would bring more invariance to learned representations but inevitably reduces the training efficiency. In Table \ref{tab:aug}, we study the influence of $N$ by gradually increasing it. It is worth mentioning that $N$=0 means the anchor view is the same as the positive view, which leads to the worst result in Table \ref{tab:aug}. The reason behind is GraVIS can easily distinguish positive views from negative ones, reducing the discriminative ability of learned representations. When $N$ is equal to 2, GraVIS degenerates into a similar format to the archetypical triplet loss but has more heterogeneous pairs. If we increase $N$ to 10, the learned invariant features provide 1.3 percents improvements over $N$=2. Obviously, as we add more augmented views, the overall segmentation performance continue to increase. However, we can find that $N$=40 is only marginally better than $N$=20, implying that the performance becomes saturated in large values of $N$. Considering the need of training efficiency (i.e., $N$=40 costs much more time and GPU memory), we set $N$ as the default value in all experiments.
\begin{table}[]
    \centering
    \caption{Investigation of $N$-times augmentation.}
    \renewcommand{\arraystretch}{1.3}
    \setlength{\abovecaptionskip}{10pt}%
    \begin{tabular}{c|c|c|c|c|c}
    \toprule
         $N$&0&2&10&20&40\\
         \hline
         Jaccard index(\%)& 52.3 &73.5&74.8&75.7&75.8\\
    \bottomrule
    \end{tabular}
    \label{tab:aug}
\end{table}

\subsubsection{Influence of the batch size $B$} Different from $N$, the batch size $B$ only influences the number of heterogeneous pairs in each loss computation. From Table \ref{tab:batch} where we gradually increase the batch size from 16 to 128, we observe an obvious trend that is increasing the input batch size would not affect as much as increasing $N$. Thus, we can draw a conclusion that the number of heterogeneous pairs play a less important role in GraVIS compared to that of homogeneous pairs. These phenomena imply that $B$=16 already provides enough negative views (i.e., 16$\times$(20-1)=304) for contrasting with the positive ones. In addition, the performance become saturated as we increase $B$ to 128. Considering large batch sizes would consume too much more GPU memory for loss computation, we set $B$ to 32 in all experiments.

\begin{table}[htp]
    \centering
    \caption{Investigation of the batch size $B$. $N$ is set to 20.}
    \renewcommand{\arraystretch}{1.3}
    \setlength{\abovecaptionskip}{10pt}%
    \begin{tabular}{c|c|c|c|c}
    \toprule
         $B$&16&32&64&128\\
         \hline
         Jaccard index(\%)&75.4&75.7&75.7&\textbf{75.8}\\
    \bottomrule
    \end{tabular}
    \label{tab:batch}
\end{table}

\begin{table}[htp]
    \centering
    \caption{Computational overhead of GraVIS under different $N$. We report the average training time (in second) per epoch.}
    \begin{tabular}{c|c|c|c|c|c}
    \toprule
         $N$ & 0 & 2 & 10 & 20 \\
         \hline
         SimCLR-Derm & - & 32 & - & - & - \\
         \hline
         GraVIS (w/o $\gamma$) & 25 & 47 & 76 & 115 \\
         GraVIS (w/ $\gamma$) & 27 & 51 & 83 & 125 \\
    \bottomrule
    \end{tabular}
    \label{comp_overhead}
\end{table}

\begin{table}[htp]
    \centering
    \caption{Investigation of the sensitivity to different augmentation strategies. \textbf{RandCrop}, \textbf{RandFlip}, \textbf{RandRot}, \textbf{ColorDist} stand for random crop, random horizontal flip, random rotation and color distortion, respectively.}
    \resizebox{1.0\columnwidth}{!}{
    \begin{tabular}{c|c|c|c|c|c}
    \toprule
         & w/o RandCrop & w/o RandFlip & w/o RandRot & w/o ColorDist & Optimal \\
         \hline
         GraVIS & 70.5 & 74.5 & 74.8 & 72.8 & 75.7\\
    \bottomrule
    \end{tabular}}
    \label{tab:crop}
\end{table}

\begin{table}[htp]
    \centering
    \caption{Investigation of the influence of $\tau$.}
    \renewcommand{\arraystretch}{1.3}
    \setlength{\abovecaptionskip}{10pt}%
    \begin{tabular}{c|c|c|c|c|c}
    \toprule
    $\tau$&0.01&0.1&0.2&0.5&1.0\\
    \hline
    Jaccard index(\%)&73.5&75.4&\textbf{75.7}&75.5&74.3\\
    \bottomrule
    \end{tabular}
    \label{tab:temp}
\end{table}

\begin{figure}[t]
\centering
\subfigure[$\tau=0.01$]{
\label{Fig.sub.1}
\includegraphics[width=0.3\columnwidth]{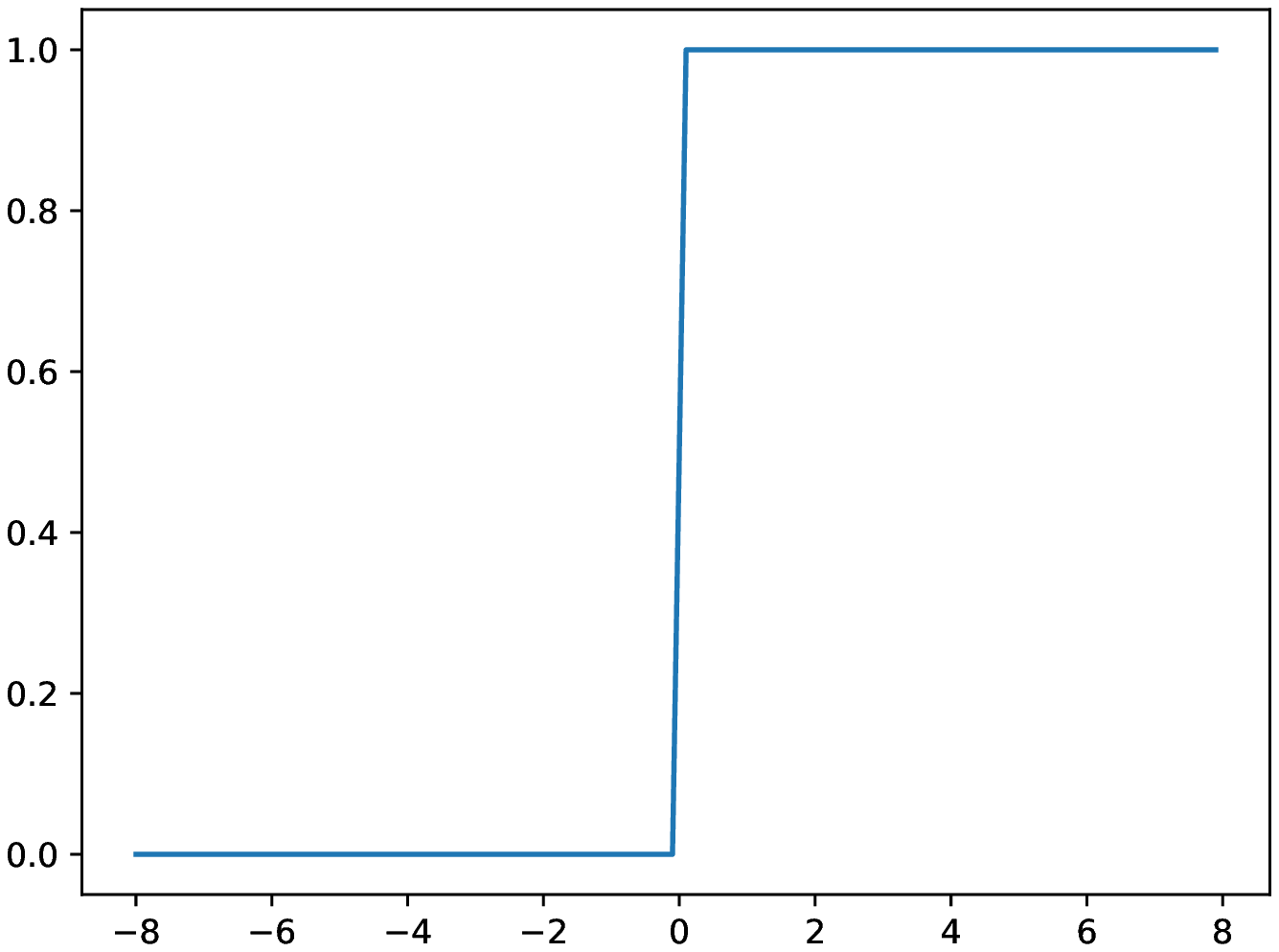}}
\subfigure[$\tau=0.2$]{
\label{Fig.sub.3}
\includegraphics[width=0.3\columnwidth]{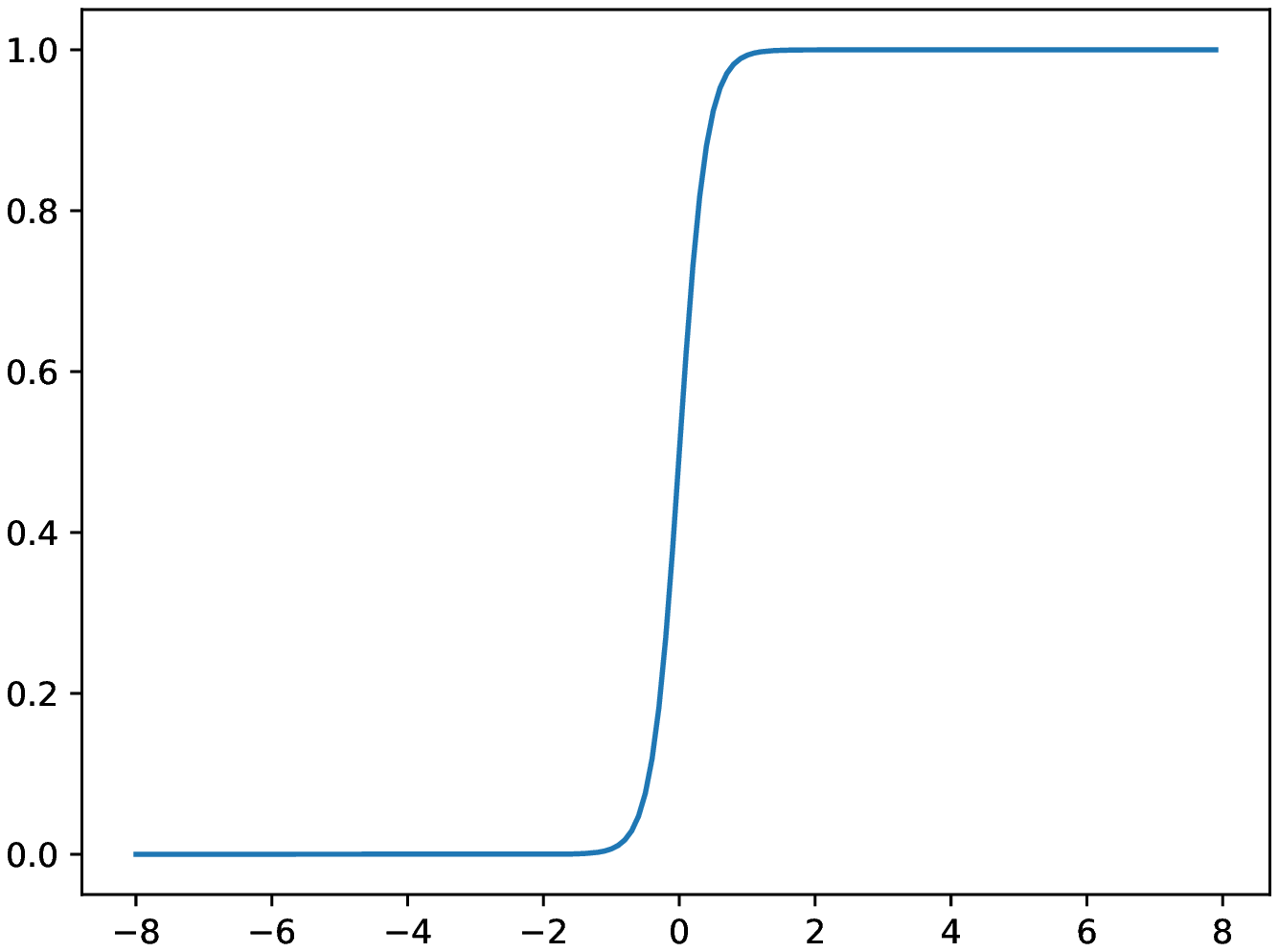}}
\subfigure[$\tau=0.5$]{
\label{Fig.sub.4}
\includegraphics[width=0.3\columnwidth]{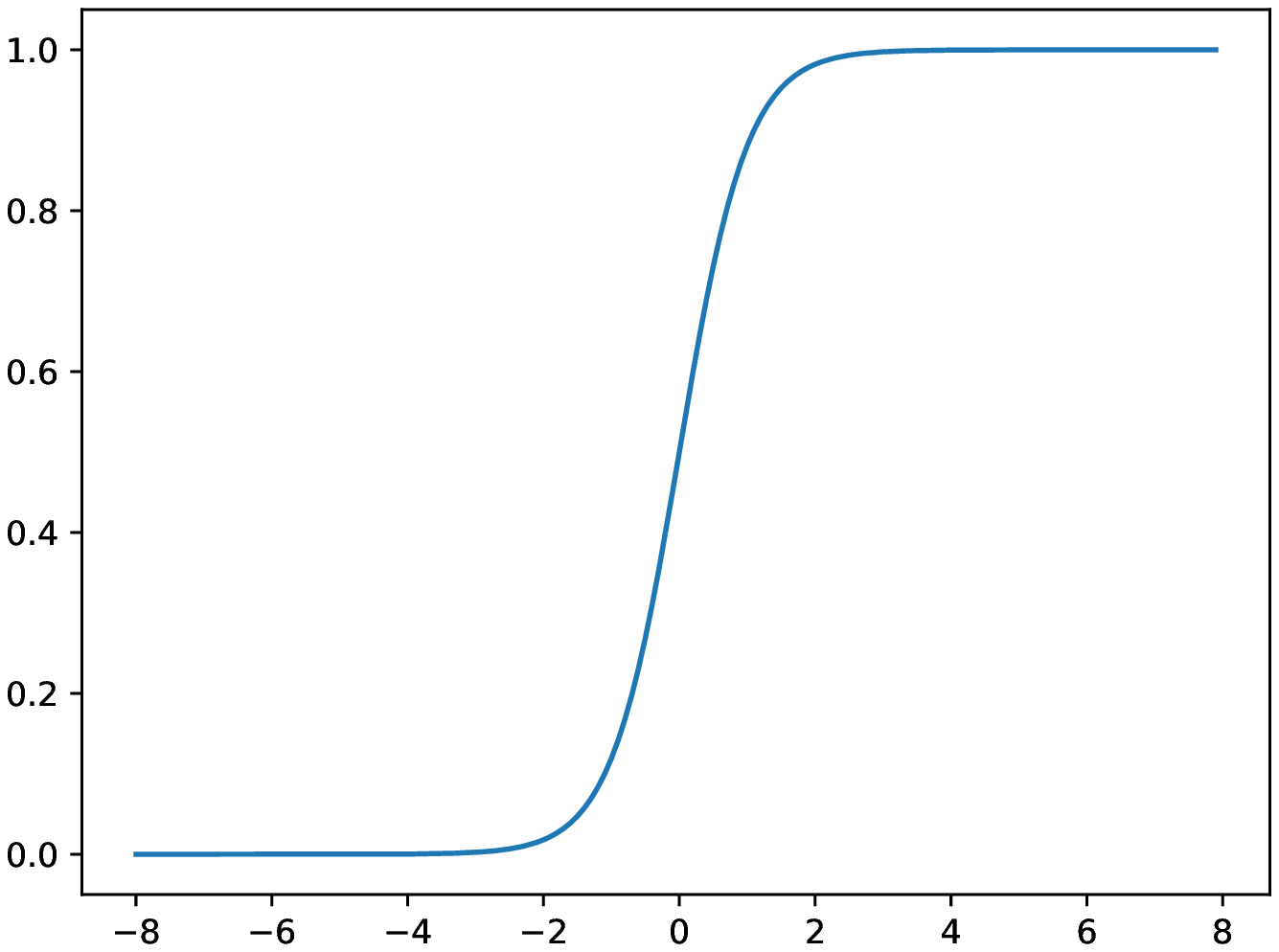}}
\caption{Sigmoid functions with different temperature values ($\tau$).}
\label{Fig.tau}
\end{figure}

\begin{figure}[t]
\centering
\subfigure[Impact of $\tau$ on the gradient]{
\label{tau_grad}
\includegraphics[width=0.48\columnwidth]{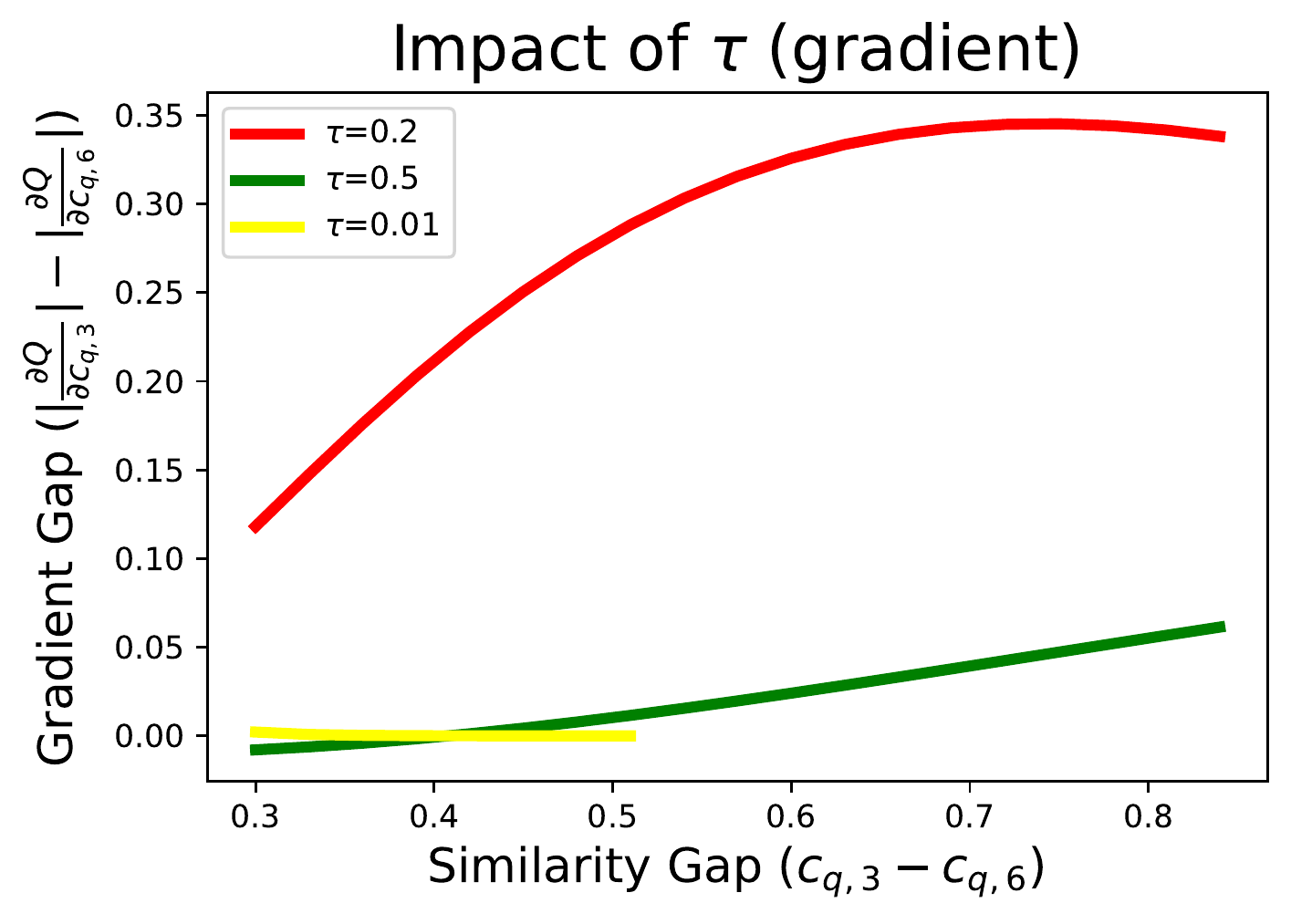}}
\subfigure[Influence of using $\sigma$ in VGL]{
\label{imp_sig}
\includegraphics[width=0.48\columnwidth]{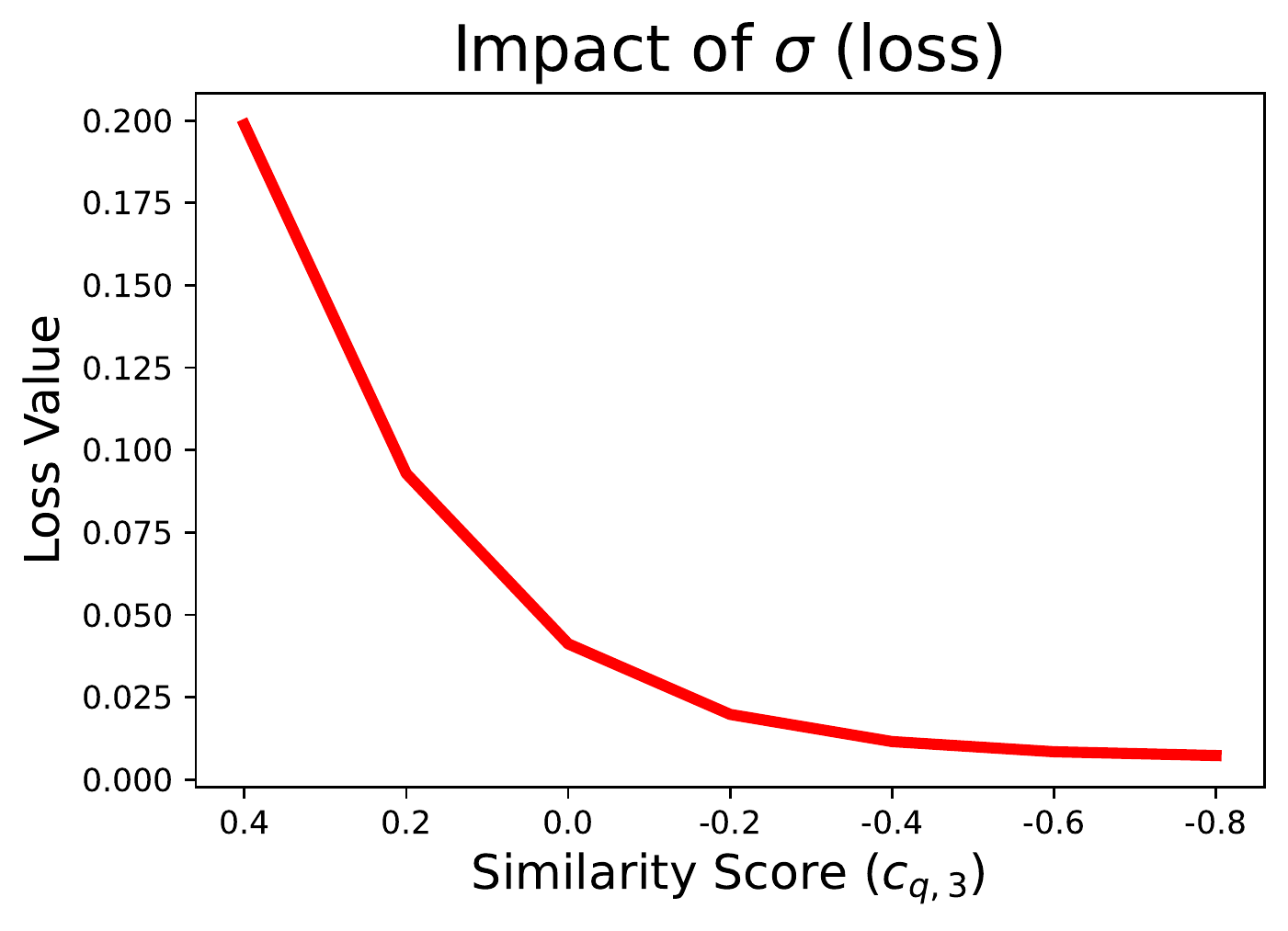}}
\caption{Impact of $\tau$ on the gradient update and investigation of the influence of using the sigmoid function $\sigma$ in VGL.}
\label{imp_tau_sig}
\end{figure}

\subsubsection{Computational overhead}In Table~\ref{comp_overhead}, we present the computational overhead of GraVIS. Besides, we also investigate the impact of adding the hardness-aware attention $\gamma$. Compared to SimCLR-Derm, our GraVIS naturally takes more training time because it includes more homogeneous and heterogeneous pairs in the view grouping loss as $N$ increases. When $N$ becomes 20 (the optimal choice), it costs about 2 minutes to finish each epoch, which is approximate 4$\times$ of SimCLR-Derm. On the other hand, we found that the hardness-aware attention $\gamma$ has little influence on the training time per epoch. Considering the obvious performance gains provided by $\gamma$ (as shown in Table~\ref{tab:att}), the helpfulness and efficiency of adding $\gamma$ to the loss function can be verified.

\subsubsection{Sensitivity to different augmentation strategies}We study the sensitivity to different augmentation strategies in Table~\ref{tab:crop}. We found that removing the random crop operation has the largest influence on the overall performance, where the jaccard index falls by over 5 percents. The reason behind is that applying random crop can dramatically increase the diversity of both homogeneous and heterogeneous image pairs. Besides random crop, it seems that color distortion becomes the most influential augmentation strategy. Adding the color distortion operation to GraVIS brings about 3-percent gains in lesion segmentation. The underlying reason is that lesions usually vary in colors, where applying color distortion helps learn color-invariant representations.

\subsubsection{Slope controller $\tau$}The temperature hyper-parameter $\tau$ in Eq. \ref{equation.sigmoid} controls the slope of the sigmoid function, which is closely related to how much GraVIS addresses similar image pairs. To achieve satisfactory performance, the sigmoid function should appropriately address image pairs with different degrees of similarity. From Fig. \ref{Fig.tau}, we can see that $\tau$ determines the range of scope (centered around zero in the horizontal axis) with large slopes. When we set $\tau$ to 0.01, the scope is narrowed. If we set $\tau$ to 0.5, the scope becomes wider. As shown in Eq. \ref{equation.sigmoid}, such scope corresponds to the similarity between two image pairs. In practice, large slopes within the scope usually lead to large gradient updates, which means GraVIS focuses more on image pairs within a specific range of similarity. Inspired by these phenomena, we study the influence of $\tau$ and report the experimental results in Table \ref{tab:temp}. It is obvious that either a too large or too small $\tau$ would adversely affect the performance of GraVIS. A too large $\tau$ (such as 0.5) cannot well address similar image pairs as the corresponding gradient updates are not prominent enough. In contrast, a too small $\tau$ (such as 0.01) only addresses very similar pairs but ignores less similar pairs. Consider $\tau$=0.2 produces the best segmentation result, we make it the default choice for $\tau$ in all experiments. 

To help explain these phenomena, we study the impact of $\tau$ on the gradient update in Fig.~\ref{tau_grad}. Here, we still refer to the example in Eq.~\ref{example}. We can see that when $\tau$ is equal to 0.2,  $|\frac{\partial Q}{\partial c_{q,3}}|$ becomes much larger than $|\frac{\partial Q}{\partial c_{q,6}}|$ as the similarity gap increases. This observation verifies that VGL can appropriately address more on the more similar homogenenous pair ($\textbf{v}_{q}, \textbf{v}^{\text{ps}}_3$) with a higher similarity score $c_{q,3}$ than the dissimilar homogeneous pair ($\textbf{v}_{q}, \textbf{v}^{\text{ps}}_6$). Particularly, when $\tau$ is equal to 0.01, we found that the gradient diminishes as the similarity gap becomes larger, which would lead to instability during training. In Fig.~\ref{imp_sig}, we investigate the influence of using the sigmoid function $\sigma$ in VGL. We found that as the similarity of a give image pair gets lower, its influence to the loss (i.e., VGL) becomes weaker. This phenomenon shows that the sigmoid function $\sigma$ may serve as a soft margin that helps prevent the effect of null similarities.

\subsection{Visual analysis}
\begin{figure}
\centering
\includegraphics[width=1.0\columnwidth]{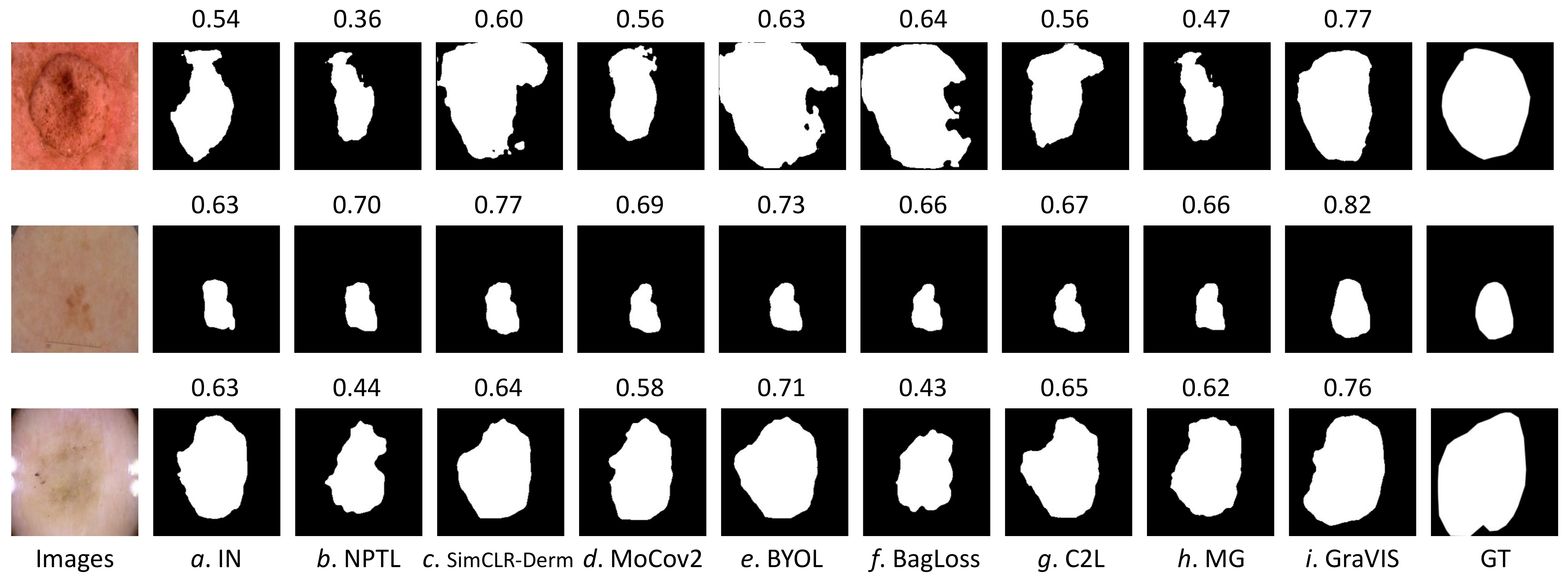}
\caption{Lesion segmentation results of different approaches. Each row includes one randomly selected test case. \textbf{GT} denotes the ground truth.}
\label{Fig.visual}
\end{figure}
In Fig. \ref{Fig.visual}, we visualize the segmentation results of different segmentation methods. The top two rows display the segmentation results of nevus, which is a non-specific medical term for a visible, circumscribed, chronic lesion of the skin or mucosa. Clinically, nevus in individuals are usually uniform in color and border. We can see that our GraVIS has the ability to detect most parts of nevus while delineating their boundary better than other approaches. In the third row, we present predicted masks of seborrheic keratosis, a non-cancerous (benign) skin tumour that originates from cells in the outer layer of the skin. Lesions of seborrheic keratosis often appear in various colors, sizes and shapes, which usually come in association with other skin conditions. Even so, GraVIS is still able to capture the majority of the main lesion site, which demonstrates the discrimination ability of learned invariant representations.

\subsection{Applying GraVIS to multiple views}
\begin{table}[t]
    \centering
    \caption{Linear fine-tuning AUCs across the tasks of Atelectasis, Cardiomegaly, Consolidation, and Edema classification on CheXpert. Note that MiCLE and GraVIS share the same augmentation strategies as used in MedAug.}
    \resizebox{1.0\columnwidth}{!}{
    \begin{tabular}{c|c|c|c|c|c}
         \toprule
         Methods & Average & Atelectasis & Cardiomegaly & Consolidation & Edema\\
         \hline
         \hline
         MedAug & 0.792 & 0.721 & 0.779 & 0.801 & 0.866 \\
         MiCLE & 0.795 & 0.728 & 0.777 & 0.794 & 0.882 \\
         \hline
         GraVIS & 0.805 & 0.735 & 0.782 & 0.824 & 0.877 \\
         \bottomrule
    \end{tabular}}
    \label{ex_mv}
\end{table}
In medical image analysis, it is possible that each patient study contains multiple views. In this case, it is intuitive to treat these views and their augmented images as positive views. In this part, we apply GraVIS to multiple views in order to investigate whether GraVIS has the ability to handle this situation. We conducted experiments on CheXpert~\cite{irvin2019chexpert}. The dataset consists of 224,316 images from 65,240 patients labeled for the presence or absence of 14 radiological observations. We use these images for pre-training and random samples of 1\% of these images for fine-tuning. We compare GraVIS with MiCLE~\cite{azizi2021big} and MedAug~\cite{vu2021medaug}, both of which were initially designed for dealing with multiple views.

In this scenario, we randomly select a view from a given patient as the anchor view, while the rest views and their augmented versions are considered as positive views. In contrast, negative views consist of chest X-rays from different patients and their augmented versions. We employed the same set of augmentation strategies as used in \cite{vu2021medaug}. For other experimental settings, we simply followed our experiences on dermatology images and report the results in Table~\ref{ex_mv}. From Table~\ref{ex_mv}, we can see that GraVIS outperforms both MedAug and MiCLE across different tasks on X-ray images, which verify the effectiveness of GraVIS in the multi-instance scenario.

\section{Discussion}
Compared to contrastive SSL approaches (i.e., NCE-based methods), GraVIS focuses more on extracting invariant semantic information given a number of homogeneous views. To address the importance of similar homogeneous pairs and prevent dissimilar homogeneous pairs from dominating the training process, a hardness-aware attention mechanism is introduced to amplify the influence of similar homogeneous pairs by assigning large penalty to them.

The main idea of GraVIS, i.e., focusing on similar homogeneous pairs to learn invariant representations, is also addressed in some recent works \cite{grill2020bootstrap,chen2021exploring} that attempt to utilize only homogeneous pairs for self-supervised representation learning. However, these approaches only display comparable results with NCE-based methods that are surpassed by our GraVIS. The improvements of GraVIS can be attributed to the fact that we focus more on homogeneous pairs while not completely ignoring heterogeneous ones (unlike \cite{grill2020bootstrap,chen2021exploring}). Thus, we believe how to appropriately address and balance the importance of both homogeneous and heterogeneous pairs can be an important direction for contrastive or non-contrastive SSL approaches.

For future work, we will explore conducting more experiments using GraVIS on a larges number of dermatology images (without labels) and fine-tune the pre-trained models on some relatively small skin datasets. We will introduce the hard triplet mining into GraVIS to see if it could help the view grouping cost to utilize heterogeneous samples as effectively as homogeneous ones. Last but not least, efforts will be made to explore a more efficient training strategy to reduce the batch size in the pre-training stage. In fact, although the batch size may not be problem in 2D medical image processing, it will become an inescapable problem in 3D medical image segmentation, which often consumes a great amount of GPU memory.

\section{Conclusion} 
We propose GraVIS to learn transferable representations in the pre-training stage by aggregating multiple homogeneous views while separating heterogeneous ones. GraVIS does not need tens of thousands of heterogeneous samples but achieves better performance than state-of-the-art contrastive pre-training methods. We conduct extensive experiments and comprehensive ablation studies to demonstrate the effectiveness of GraVIS. We found that when pre-trained with GraVIS, a single model can already achieve better results compared to the winners and recent state-of-the-art approaches in ISIC 2017 challenges. We believe these phenomena suggest that GraVIS has the potential to replace canonical NCE-based SSL approaches.

{
\bibliographystyle{ieeetr}
\bibliography{egbib}
}
\end{document}